\documentclass[11pt]{article}
\usepackage[english]{babel}
\usepackage{a4wide,colacl,qtree}

\begin{document}

\title{Aspects of Pattern-matching in Data-Oriented Parsing}
\author{Guy De Pauw
\\ CNTS - Language Technology Group
\\ {\small UIA - University of Antwerp}
\\ {\small Universiteitsplein 1, 2610 Antwerpen, Belgium}
\\ depauw@uia.ua.ac.be}

\maketitle

\begin{abstract}
{\footnotesize Data-Oriented Parsing ({\sc dop}) ranks among the best parsing schemes, pairing state-of-the art
parsing accuracy to the psycholinguistic insight that larger chunks of syntactic structures are relevant grammatical
and probabilistic units. Parsing with the {\sc dop}-model, however, seems to involve a lot of CPU cycles and a considerable
amount of double work, brought on by the concept of multiple derivations, which is necessary for probabilistic
processing, but which is not convincingly related to a proper linguistic backbone. It is however possible to
re-interpret the {\sc dop}-model as a pattern-matching model, which tries to maximize the size of the substructures that
construct the parse, rather than the probability of the parse. By emphasizing this memory-based aspect of the {\sc dop}-model,
it is possible to do away with multiple derivations, opening up possibilities for efficient Viterbi-style
optimizations, while still retaining acceptable parsing accuracy through enhanced context-sensitivity.}
\end{abstract}

\begin{figure*}
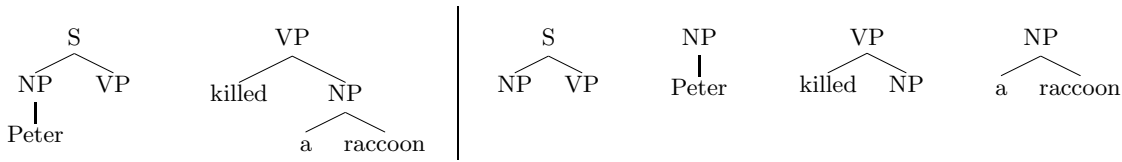

{\footnotesize
\qtreecenterfalse
\begin{tabular}{ll|llll}
\Tree [.S [.NP Peter ] VP ] & \Tree [.VP killed [.NP a raccoon ] ] &  \Tree [.S NP VP ] & \Tree [.NP Peter ] & \Tree
[.VP killed NP ] & \Tree [.NP a raccoon ]\\
\end{tabular}}
\caption{Multiple Derivations} \label{deriv}
\end{figure*}

\section{Introduction}

The machine learning paradigm of Memory-Based Learning, based on the assumption that new problems are solved by direct
reference to stored experiences of previously solved problems, has been successfully applied  to a number of linguistic
phenomena, such as part-of-speech tagging, NP-chunking and stress acquisition (consult \newcite{daelemans99} for an overview).
To solve these particular problems, linguistic information needed to trigger the correct
disambiguation, is encoded in a linear {\em feature value} representation and presented to a memory based learner,
such as TiMBL \cite{Daelemans+99b}.

Yet, many of the intricacies of the domain of syntax do not translate well to a linear representation, so that
established MBL-methods are necessarily limited to low-level syntactic analysis, like the aforementioned NP-chunking
task.

Data Oriented Parsing \cite{bod:1999a}, a state-of-the art natural language parsing system, translates very well to a
Memory Based Learning context. This paper describes a re-interpretation of the {\sc dop}-model, in which the {\em
pattern-matching aspects} of the model are exploited, so that parses are analyzed by trying to match a new analysis to
the largest possible substructures recorded in memory.

A short introduction to Data Oriented Parsing will be presented in Section \ref{dop}, followed by an explanation of the term {\em
pattern-matching} in the context of this paper. Section \ref{setup} describes the experimental setup and the corpus. The parsing
phase that precedes the disambiguation phase will be outlined in Section \ref{parsing} and a description of the 3 disambiguating models, {\sc pcfg}, {\sc pmpg} and the combined system {\sc pcfg+pmpg} can  be found in Sections \ref{pcfg}, \ref{pmpg} and \ref{combi}.

\section{Data Oriented Parsing}
\label{dop}
Data Oriented Parsing, originally conceived by Remko Scha \cite{scha90}, has been successfully applied to syntactic
natural language parsing by Rens Bod \shortcite{Bod95}, \shortcite{bod:1999a}. The aim of Data Oriented Parsing (henceforth {\sc dop}) is to
develop a {\em performance model} of natural language, that models language use rather than some type of
competence. It adapts the psycholinguistic insight that language users analyze sentences using previously registered
constructions and that not only rewrite rules, but complete substructures of any given depth can be linguistically relevant
units for parsing.

\subsection{Architecture}
The core of a {\sc dop}-system is its {\sc treebank}: an annotated corpus is used to induce all substructures of
arbitrary depth, together with their respective probabilities, which is a expressed by its frequency in the
{\sc treebank} relative to the number of substructures with the same root-node.

Figure \ref{deriv} shows the combination operation that is needed to form the correct parse tree for the sentence {\em Peter
killed a raccoon}. Given a treebank of substructures, the system tries to match the leftmost open node of a
substructure that is consistent with the parse tree, with the top-node of another substructure, consistent with the
parse tree.

Usually, different combinations of substructures are possible, as is indicated in Figure \ref{deriv}: in the
example at the left-hand side the tree-structure can be built by combining an {\sc s}-structure with a specified 
{\sc np} and a fully specified {\sc vp}-structure. The right example shows another possible combination, where a parse 
tree is built by combining the minimal substructures. Note that these are consistent with ordinary rewrite-rules, such 
as {\sc s \begin{math} \rightarrow \end{math} np vp}.

One particular parse tree may thus consist of several different {\em derivations}. To find the probability of a
derivation, we multiply the probabilities of the substructures that were used to form the derivation. To find the
probability of a parse, we must in principle sum the probabilities of all its derivations.

It is computationally hardly tractable to consider all derivations for each parse. Since {\sc viterbi} optimization
only succeeds in finding the most probable derivation as opposed to the most probable parse, the {\sc monte carlo}
algorithm is introduced as a proper approximation that randomly generates a large number of derivations. The most
probable parse is considered to be the parse that is most often observed in this derivation forest.

\subsection{Experimental Results of {\sc dop}} \label{experimental}
The basic {\sc dop}-model, {\sc dop1}, was tested on a manually edited version of the ATIS-corpus \cite{Marcus+93}. The
system was trained on 603 sentences (part-of-speech tag sequences) and evaluated on a test set of 75 sentences. Parse 
accuracy was used as an evaluation metric, expressing the percentage of sentences in the test set for which the parse 
proposed by the system is completely identical to the one in the original corpus. Different experiments were conducted
in which maximum substructure size was varied. With {\sc dop1}-limited to a substructure-size of 1 (equivalent to a
{\sc pcfg}), parse accuracy is 47\%. In the optimal {\sc dop}-model, in which substructure-size is not limited, a parse
accuracy of \textbf{85\%} is obtained.

\subsection{Short Assessment of DOP}
{\sc dop1} in its optimal form achieves a very high parse accuarcy. The computational
costs of the system, however, are equally high. \newcite{Bod95} reported an average parse time of 3.5 hours per
sentence. Even though current parse time is reported to be more reasonable, the optimal {\sc dop} algorithm in which no
constrains are made on the size of substructures, may not yet be tractable for life-size corpora.

In a context-free grammar framework (consistent with {\sc dop} limited to a substructure-size of 1), there is only one
way a parse tree can be formed (for example, the right hand side of Figure \ref{deriv}), meaning that there is only one
derivation for a given parse tree. This allows efficient {\sc viterbi} style optimization.

To encode context-sensitivity in the system, {\sc dop} is forced to introduce multiple derivations, so that repeatedly
the same parse tree needs to be generated, bringing about a lot of computational overhead.

Even though the use of larger syntactic contexts is highly relevant from a psycholinguistic point-of-view, there is no
explicit preference being made for larger substructures in the {\sc dop} model. While the {\sc Monte Carlo}
optimization scheme maximizes the probability of the derivations and seems to prefer derivations made up of
larger substructures, it may be interesting to see if we can make this assumption explicit.

\begin{table*}
\footnotesize{
\begin{tabular}{c|cc|c||cc}\hline
    Disambiguator & Parse Accuracy (/562)   &  \%           & F    & Parse Accuracy on parsable sentences (/456)& \%\\ \hline
    {\sc pcfg}          & 373                     & \textbf{66.4}  & 83.0 & 373 & 81.8\\
    {\sc pmpg}          & 327                     & \textbf{58.2}  & 75.1 & 327 & 71.7\\
    {\sc pcfg+pmpg}     & 402                     & \textbf{71.5}  & 85.2 & 402 & 88.2\\
\end{tabular}}
\caption{Experimental Results} \label{results}
\end{table*}

\begin{figure*}
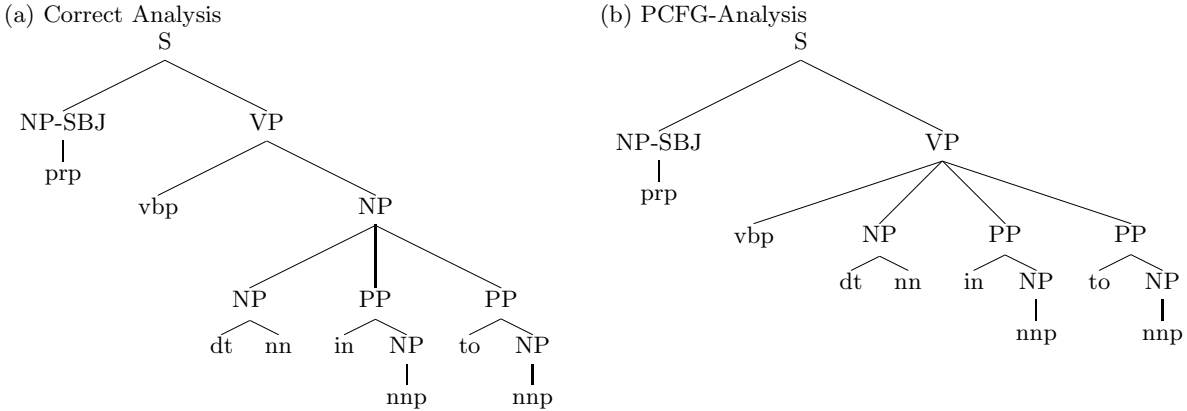

{\footnotesize
\begin{tabular}{p{7.5cm}p{8cm}}
    (a) Correct Analysis & (b) PCFG-Analysis\\
    \qtreecenterfalse
    \Tree [.S [.NP-SBJ prp ] [.VP vbp [.NP [.NP dt nn ] [.PP in [.NP nnp ] ] [.PP to [.NP nnp ] ] ] ] ] &
    \qtreecenterfalse
    \Tree [.S [.NP-SBJ prp ] [.VP vbp [.NP dt nn ] [.PP in [.NP nnp ] ] [.PP to [.NP nnp ] ] ] ] \\
\end{tabular}}
\caption{{\sc pcfg} Error Analysis} \label{PCFG-eran}
\end{figure*}

\section{Pattern-matching}
\label{pm}

When we look at natural language parsing from a memory-based point of view, one might say that a sentence is analyzed
by looking up the most similar structure for the different analyses of that sentence in memory. The parsing system 
described in this paper tries to mimic this behavior by interpreting the {\sc dop}-model as a memory-based model, in 
which analyses are being matched with syntactic {\em patterns} recorded in memory. Similarity between the proposed analysis 
and the patterns in memory is computed according to:

\begin{itemize}
\item the number of patterns needed to construct a tree (to be minimized)
\item the size of the patterns that are used to construct a tree (to be maximized)
\end{itemize}

The {\em nearest neighbor} for a given analysis can be defined as the derivation that shares the largest amount of common nodes.

\section{The experimental Setup}
\label{setup}
10-fold cross-validation was used to appropriately evaluate the algorithms, as the dataset (see Section \ref{corpus}) is
rather small. Like {\sc dop1} the system is trained and tested on part-of-speech tag sequences. In a first phase, a
simple bottom-up chart parser, trained on the training partitions, was used to generate parse forests for the
part-of-speech tag sequences of the test partition. Next, the parse forests were sent to the 3 algorithms (henceforth
the {\em disambiguators}) to order these parse forests, the first parse of the ordered parse forest being the one
proposed by the disambiguator.

In this paper, 3 disambiguators are described:

\begin{itemize}
    \item {\sc pcfg}: simple Probabilistic Context-Free Grammar
    \item {\sc pmpg}: the {\sc dop} approximation, Pattern-Matching Probabilistic Grammar
    \item {\sc pcfg+pmpg}: a combined system, integrating {\sc pcfg} and {\sc pmpg}
\end{itemize}

The evaluation metric used is parse accuracy, but also the typical parser evaluation metric {\em F-measure}
(precision/recall) is given as a means of reference to other systems.

\subsection{The Corpus}
\label{corpus} The experiments were conducted on an edited version of the ATIS-II-corpus \cite{Marcus+93}, which
consists of 578 sentences. Quite a lot of errors and inconsistencies were found, but not corrected, since we want our
(probabilistic) system to be able to deal with this kind of noise. Semantically oriented flags like {\sc -tmp} and {\sc
-dir}, most often used in conjunction with {\sc pp}, have been removed, since there is no way of retrieving this kind
of semantic information from the part-of-speech tags of the ATIS-corpus. Syntactic flags like {\sc -sbj}, on the other
hand, have been maintained. Internal relations (denoted by numeric flags) were removed and for practical reasons,
sentence-length was limited to 15 words max. The edited corpus retained 562 sentences.

\section{Parsing}
\label{parsing} As a first phase, a bottom-up chart parser parsed the test set. This proved to be quite problematic,
since overall, 106 out of 562 sentences (19\%) could not be parsed, due to the sparseness of the grammar, meaning that the
appropriate rewrite rule needed to construct the correct parse tree for a sentence in the test set, wasn't featured in
the induced grammar. {\sc np}-annotation seemed to be the main cause for unparsability. An {\sc np} like {\em
restriction code AP/57} is represented by the rewrite rule:

{\footnotesize NP \begin{math}\rightarrow\end{math} NN NN sym sym sym CD CD}
\vspace{0.2cm}

Highly specific and flat structures like these are scarce and are usually not induced from the training set when needed
to parse the test set.

On-going research tries to implement grammatical smoothing as a solution to this problem, but one might also consider
generating parse forests with an independent grammar, induced from the entire corpus (training set+testset) or a
different corpus. In both cases, however, we would need to apply probabilistic smoothing to be able to assign
probabilities to unknown structures/rules. Neither grammatical, nor probabilistic smoothing was implemented in the
context of the experiments, described in this paper.

The sparseness of the grammar proves to be a serious bottleneck for parse accuracy, limiting our disambiguators to a
maximum parse accuracy of 81\%.

\section{{\sc pcfg}-experiments}
\label{pcfg} A {\sc pcfg} constructs parse trees by using simple rewrite-rules. The probability of a parse tree can be
computed by multiplying the probabilities of the rewrite-rules that were used to construct the parse. Note that a {\sc
pcfg} is identical to {\sc dop1} when we limit the maximum substructures size to 1, only allowing derivations of the
type found at the right-hand side of Figure \ref{deriv}.

\subsection{Experimental Results}
The first line of Table \ref{results} shows the results for the {\sc pcfg}-experiments: 66.4\% parse accuracy is an
adequate result for this baseline model. We also look at parse accuracy for parsable sentences (an estimate of the
parse accuracy we might get if we had a more suited parse forest generator) and we notice that we are able to achieve a
81.8\% parse accuracy. This is already quite high, but on examining the parsed data, serious and fundamental
limitations to the {\sc pcfg}-model can be observed

\subsection{Error Analysis}

Figure \ref{PCFG-eran}, displays the most common type of mistake made by {\sc pcfg}'s. The correct parse tree could
represent an analysis for the sentence:

\vspace{0.2cm} {\footnotesize {\em I want a flight from Brussels to Toronto.}} \vspace{0.2cm}

This example shows that a {\sc pcfg} has a tendency to prefer flatter structures over embedded structures. This is a
trivial effect of the mathematical formula used to compute the probability of a parse-tree: embedded structure require
more rewrite rules, adding more factors to the multiplication, which will almost inevitably result in a lower
probability.

It is an unfortunate property of {\sc pcfg}'s that the number of nodes in the parse tree is inversely proportionate to
its probability. One might be inclined to normalize a parse tree's probability relative to the number of nodes in the
tree, but a more linguistically sound alternative is at hand: the enhancement of context sensitivity through the use of
larger syntactic context within parse trees can make our disambiguator more robust.

\section{{\sc pmpg}-experiments}
\label{pmpg} The Pattern-Matching Probabilistic Grammar is a memory-based interpretation of a {\sc dop}-model, in which
a sentence is analyzed by matching the largest possible chunks of syntactic structure on the sentence. To compile parse trees
into patterns, all substructures in the training set are encoded by assigning them specific indexes, {\sc np@345} e.g. denoting a fully specified {\sc
np}-structure. This approach was inspired by \newcite{goodman:1996}, in which Goodman unsuccessfully uses a system of indexed
parse trees to transform {\sc dop} into an equivalent {\sc pcfg}. The system of indexing (which is detailed in \newcite{depauw:2000a}) used
in the experiments described in this paper, is however specifically geared towards encoding contextual information in parse trees.

Given an indexed training set, indexes can then be matched on a test set parse tree in a bottom-up fashion. In the
following example, boxed nodes indicate nodes that have been retrieved from memory.

{\footnotesize
\qtreecenterfalse \Tree [.S [.\fbox{NP-SBJ} prp ] [.VP vbp [.\fbox{NP} [.\fbox{NP} dt nn ] [.\fbox{PP}
in [.\fbox{NP} nnp ] ] ] ] ]}

In this example we can see that an {\sc np}, consisting of a fully specified embedded {\sc np} and {\sc pp}, has been completely
retrieved from memory, meaning that the {\sc np} in its entirety can be observed in the training set.  However, no {\sc vp} 
was found that consists of a {\sc vbp} and that particular {\sc np}. Disambiguating with {\sc pmpg} consequently involves 
pruning all nodes retrieved from memory:

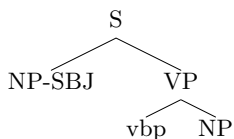
\qtreecenterfalse {\footnotesize \Tree [.S NP-SBJ [.VP vbp NP ] ]}

Finally, the probability for this pruned parse tree is computed in a {\sc pcfg}-type manner, not adding the retrieved
nodes to the product:

\vspace{0.2cm} {\footnotesize
    P(parse) = P({\sc s} \begin{math}\rightarrow\end{math} {\sc np-sbj vp}) . P({\sc vp} \begin{math}\rightarrow\end{math}
     vbp {\sc np})}
\vspace{0.2cm}

\subsection{Experimental Results}
The results for the {\sc pmpg}-experiments can be found on the second line of Table \ref{results}. On some partitions,
{\sc pmpg} performed insignificantly better than {\sc pcfg}, but Table \ref{results} shows that the results for the
context sensitive scheme are much worse. 58.2\% overall parse accuracy and 71.7\% parse accuracy on parsable sentences
indicates that {\sc pmpg} is not a valid approximation of {\sc dop}'s context-sensitivity.

\subsection{Error Analysis}
The dramatic drop in parsing accuracy calls for an error analysis of the parsed data. Figure \ref{pmpg-eran} is a
prototypical mistake {\sc pmpg} has made. The correct analysis could represent a parse tree for a sentence like:

\vspace{0.2cm} {\footnotesize {\em What flights can I get from Brussels to Toronto.}} \vspace{0.2cm}

 The {\sc pmpg} analysis would never have been considered a likely candidate
by a common {\sc pcfg}. This particular sentence in fact was effortlessly disambiguated by the {\sc pcfg}
. Yet the fact
that large chunks of tree-structure are retrieved from memory, make it the preferred parse for the {\sc pmpg}. We
notice for instance that a large part of the sentence can be matched on an {\sc sbar} structure, which has no relevance
whatsoever.

Clearly, {\sc pmpg} overestimates substructure size as a feature for disambiguation. It's interesting however to see
that it is a working implementation of context sensitivity, eagerly matching patterns from memory. At the same time, it
has lost track of common-sense {\sc pcfg} tactics. It is in the combination of the two that one may find a decent
disambiguator and accurate implementation of context-sensitivity.

\section{A Combined System ({\sc pmpg+pcfg})}
\label{combi}
Table \ref{results} showed that 81.8\% of the time, a {\sc pcfg} finds the correct parse (for parsable
sentences), meaning that the correct parse is at the first place in the ordered parse forest. \textbf{99\%} of the time, the
correct parse can be found among the 10 most probable parses in the ordered parse forest. This opens up a myriad of possibilities
for optimization. One might for instance use a {\em best-first} strategy to generate only the 10 best parses,
significantly reducing parse and disambiguation time. An optimized disambiguator might therefore include a preparatory
phase in which a common-sense {\sc pcfg} retains the most probable parses, so that a more sophisticated follow-up
scheme need not bother with senseless analyses.

In our experiments, we combined the common-sense logic of a {\sc pcfg} and used its output as the {\sc pmpg}'s input. 
This is a well-established technique usually referred to as {\em system combination} (see \newcite{halteren+98} for an
application of this technique to part-of-speech tagging):

{\footnotesize
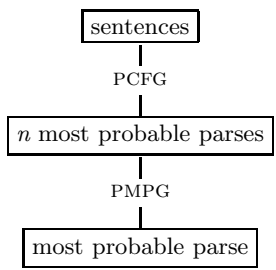
\begin{tabular}{c}
     \Tree [.\fbox{sentences} [.{\sc pcfg} [.\fbox{{\em n} most probable parses} [.{\sc pmpg} \fbox{most probable
parse} ] ] ] ]
\end{tabular}}
\vspace{0.2cm}

We are also presented with the possibility to assign a weight to each algorithm's decision. The probability of a parse
can the be described with the following formula:

\vspace{0.2cm}
\begin{math}
P(parse) = \frac{\prod\limits_{i}P(\mbox{rewrite-rule})_i}{(\mbox{\# non-indexed nodes})^n}
\end{math}
\vspace{0.2cm}

The weight of each algorithm's decision, as well as the number of most probable parses that are extrapolated for the
pattern-matching algorithm, are parameters to be optimized. Future work will include evaluation on a validation set to
retrieve the optimal values for these parameters.

\subsection{Results}
The third line in Table \ref{results} shows that the combined system performs better than either one, with a parse
accuracy of 71.5\% and close to 90\% parse accuracy on parsable sentences, which we can consider an approximation of
results reported for {\sc dop1}. Error analysis shows that the combined system is indeed able to overcome difficulties
of both algorithms. The example in Figure \ref{PCFG-eran} as well as the example in Figure \ref{pmpg-eran} were
disambiguated correctly using the combined system

\begin{figure*}
{\footnotesize
\begin{tabular}{p{14cm}}
    (a) Correct Analysis \\
    \qtreecenterfalse \Tree [.S [.WHNP [.WHNP wdt nns ] [.PP xxx ] [.PP xxx ] ] [.SQ vbp [.NP-SBJ prp ] [.VP vb [.NP xxx ]
[.PP in [.NP nnp ] ] [.PP to [.NP nnp ] ] ] ] ] \\
    (b) {\sc pmpg} Analysis\\
    \qtreecenterfalse \Tree [.S [.NP-SBJ [.\fbox{NP} wdt nns ] [.\fbox{SBAR} [.\fbox{WHNP} xxx ] [.\fbox{S} [.\fbox{NP-SBJ} xxx ] [.\fbox{VP} vb [.\fbox{NP} prp ] ] ] ]
    ]
                            [.\fbox{VP} vb [.\fbox{NP} xxx ] [.\fbox{PP} in [.\fbox{NP} nnp ] ] [.\fbox{PP} to [.\fbox{NP} nnp ] ] ]]
\faketreewidth{www}

\end{tabular}}
\caption{{\sc pmpg} Error Analysis} \label{pmpg-eran}
\end{figure*}

\section{Future Research}
Even though the {\sc pmpg} shows a lot of promise in its parse accuracy, the following extensions need to be
researched:

\begin{itemize}
    \item Optimizing {\sc pmpg+pcfg} for computational efficiency: the graph in Section \ref{combi} shows a possible
    optimized parsing system, in which a pre-processing {\sc pcfg} generates the {\em n} most likely candidates to be
    extrapolated for the actual disambiguator. Full parse forests were generated for the experiments described in this
    paper, so that the efficiency gain of such a system cannot be properly estimated.
    \item {\sc pmpg+pcfg} as an approximation needs to be compared to actual {\sc dop}, by having {\sc dop} parse the data used
    in this experiment, and by having {\sc pmpg+pcfg} parse the data used in the experiments described in \newcite{bod:1999a}.
    \item The bottleneck of the sparse grammar problem prevents us from fully exploiting the disambiguating power of the
    pattern-matching algorithm. The {\sc grael}-system (GRammar Adaptation, Evolution and Learning) that is currently being
    developed, tries to address the problem of grammatical sparseness by using evolutionary techniques to generate,
    optimize and complement grammars.
\end{itemize}

\section{Conclusions}
Even though {\sc dop1} exhibits outstanding parsing behavior, the efficiency of the model is rather problematic. The
introduction of multiple derivations causes a considerable amount of computational overhead. Neither is it clear how
the concept of multiple derivations translates to a psycholinguistic context: there is no proof that language users
consider different instantiations of the same parse, when deciding on the correct analysis for a given sentence.

A pattern-matching scheme was presented that tried to disambiguate parse forests by trying to maximize the size of the
substructures that can be retrieved from memory. This straightforward memory-based interpretation yields sub-standard
parsing accuracy. But the combination of common-sense probabilities and enhanced context-sensitivity provides a
workable parse forest disambiguator, indicating that language users might exert a complex combination of memory-based
recollection techniques and stored statistical data to analyze utterances.

\end{document}